# SyMPox: An Automated Monkeypox Detection System Based on Symptoms Using XGBoost

Alireza Farzipour[1], Roya Elmi[2], Hamid Nasiri[3,*]

[1] Department of Computer Science, Semnan University, Semnan, Iran.
[2] Farzanegan Campus, Semnan University, Semnan, Iran.
[3] Department of Computer Engineering, Amirkabir University of Technology (Tehran Polytechnic), Tehran, Iran.

Correspondence should be addressed to Hamid Nasiri; h.nasiri@aut.ac.ir

**Abstract**

Monkeypox is a zoonotic disease. About 87000 cases of monkeypox were confirmed by the World Health Organization until 10th June 2023. The most prevalent methods for identifying this disease are image-based recognition techniques. Still, they are not too fast and could only be available to a few individuals. This study presents an independent application named SyMPox, developed to diagnose Monkeypox cases based on symptoms. SyMPox utilizes the robust XGBoost algorithm to analyze symptom patterns and provide accurate assessments. Developed using the Gradio framework, SyMPox offers a user-friendly platform for individuals to assess their symptoms and obtain reliable Monkeypox diagnoses.

**Code metadata**

| Nr | Code metadata description | |
|---|---|---|
| C1 | Current code version | *V1.0* |
| C2 | Permanent link to code/repository used for this code version | *https://github.com/alirezafarzipour/SyMPox* |
| C3 | Permanent link to reproducible capsule | *https://codeocean.com/capsule/0867664/tree/v1* |
| C4 | Legal code license | *GNU General Public License v3.0* |
| C5 | Code versioning system used | *git* |
| C6 | Software code languages used | *Python* |
| C7 | Compilation requirements, operating environments and dependencies | *Python 3.8 or later*<br>*Pandas, NumPy, Gradio, Scikit-learn, XGboost, imblearn* |
| C8 | If available, link to developer documentation/manual | *-* |
| C9 | Support email for questions | *h.nasiri@aut.ac.ir* |

**Software metadata**

| | |
|---|---|
| Current software version | *1.5.3* |
| Permanent link to executables of this version | *https://github.com/alirezafarzipour/SyMPox* |
| Legal Software License | *GNU General Public License v3.0* |
| Operating System | *Microsoft Windows 7 (or later)*<br>*Mac OS 10.12.6 (Sierra or later)*<br>*Linux* |
| Installation requirements & dependencies | *4 GB of memory*<br>*1 GB of free disk space* |

**1. Introduction**

Monkeypox is a zoonotic viral infection that manifests symptoms similar to those observed in individuals with smallpox. In contrast to the transmission of smallpox and chickenpox viruses, which primarily occur through direct person-to-person contact, the Monkeypox virus (MPXV) can be transmitted between animals and humans through blood and other bodily fluids [1]. Monkeypox is not an emerging or newly discovered disease [2]. The initial incidence of the infection was detected as early as 1970, with subsequent cases exhibiting an upward trend over the ensuing decade. The PCR (Polymerase Chain Reaction) test is utilized to detect Monkeypox. Nevertheless, because the virus may only stay in the blood for a brief period, the findings of this test may not be reliable and conclusive [3], [4].

Since monkeypox affects and manifests on human skin, it can be discerned through skin imaging. Most of the performed works can only detect monkeypox from the skin image. Jaradat et al. [5] found MobileNetV2 provided the best outcome with an accuracy of 0.98 on different datasets. Altun et al. employed the same approach based on deep learning [6]. Khafaga et al. [7] used a deep convolutional neural network to diagnose monkeypox with 0.98 accuracy. However, the accuracy of these methods is limited as additional symptoms, including headache, fever, muscle aches, back pain, and low energy, accompany the condition. Consequently, reliance solely on the image is insufficient, and due consideration must always be given to the associated symptoms. This study primarily emphasizes the diagnosis of monkeypox disease based on symptom analysis. The symptoms are examined using the XGBoost algorithm, determining the diagnostic outcome. The SMOTE (Synthetic Minority Oversampling Technique) method was employed to avoid imbalances in the dataset.

## 2. Dataset

The study utilized a dataset obtained from Kaggle by ''Larxel" titled "Global Monkeypox Cases (daily updated)" [8], consisting of global monkeypox cases reported by "Global Health" and used by the "World Health Organization." The dataset contained over 30 fields, but only the "Symptoms" and "Status" columns were considered for analysis. The data underwent cleaning and pre-processing steps, resulting in a final dataset of 211 identifiable cases. The dataset included information on symptoms and disease status, with 46 columns. The test set for all models was set at 20%.

## 3. Extreme Gradient Boosting

XGBoost is a distributed gradient-boosting framework designed to be efficient, adaptable, and transferable [9]–[11]. In 2016, the XGBoost algorithm was developed by Chen and Guestrin [12]. This method represents an enhanced version of the Gradient Boosted Decision Tree (GBDT) technique [13]–[15]. The system is equipped with functionalities such as an approximate greedy search [16] and hyper-parameters that serve to improve the learning process and mitigate overfitting [17]–[19].

## 4. SMOTE (Synthetic Minority Over-sampling Technique)

In 2002, Nitesh Chawla et al. [20] presented the most commonly employed method for generating novel instances. The SMOTE operates by identifying instances near the feature space, establishing a line between these instances within the feature space, and generating a new sample at a point along that line. The SMOTE can be employed to generate a sufficient number of synthetic instances for the underrepresented class. This approach works well because it can create new synthetic examples from the minority class that are believable and similar to the existing examples from the same class in terms of their features [21].

## 5. Software Features

SyMPox is built on XGBoost classifiers. The software utilizes Python and Gardio to create SyMPox using machine learning algorithms, offering optimal performance and an easily operated experience [22]. Gradio v3.35.2 and Python v3.11.3 are both used in the produced program.

Considering that the dataset was derived from the Global Health report and predominantly comprised individuals diagnosed with monkeypox, the number of samples representing healthy individuals within the obtained dataset is comparatively limited. Consequently, the dataset's inherent imbalance can significantly impact the performance of any classifier model. The approach of oversampling the minority class has been employed to tackle the issue of imbalanced datasets, commonly known as the SMOTE method. This technique involves augmenting the data for the minority class, effectively addressing the class imbalance. The workflow design for the proposed method is depicted in Figure. 1.

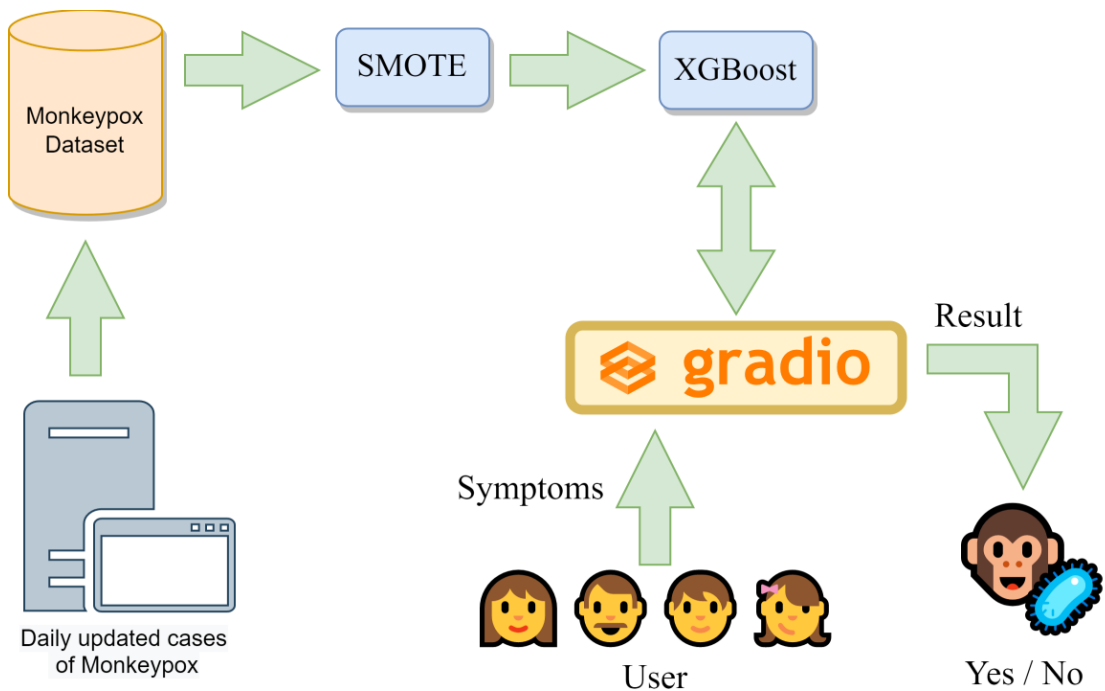

**Figure 1.** workflow of the proposed method

XGBoost was selected because it facilitates goal function customization and reduces the requisite feature analysis [23]–[25]. This implies that the model can be adjusted to enhance its efficiency in addressing the particular issue. XGBoost is known for its cost-effectiveness in computation owing to its parallel processing implementation [26]–[29]. The study used trial and error to determine the ideal algorithm's parameters for the used model, as shown in (Table. 1).

**Table 1:** XGBoost Model Parameters

| Parameter | Value |
| --- | --- |
| Base Learner | Gradient boosted tree |
| Tree construction algorithm | Exact greedy |
| Learning rate ($\eta$) | 0.0991 |
| Lagrange multiplier ($\gamma$) | 0 |
| Number of gradients boosted trees | 80 |

This software performs a single integrated process, which involves receiving the user's symptoms and classifying them to detect the presence of Monkeypox disease. During the initial phase, users input their symptoms (Figure. 2). Subsequently, the entered data undergo classification using the XGBoost algorithm. Finally, the application presents the user with the ultimate diagnosis (Figure. 3). Notably, SyMPox achieved an impressive accuracy rate of 94.64%.

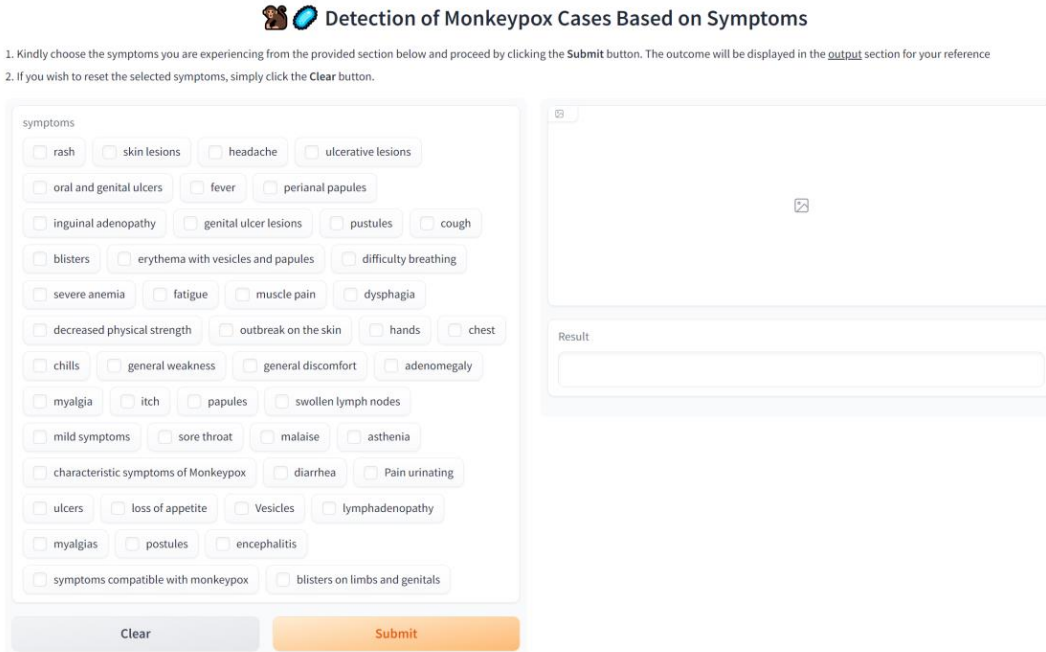

**Figure 2.** Homepage of SyMPox.

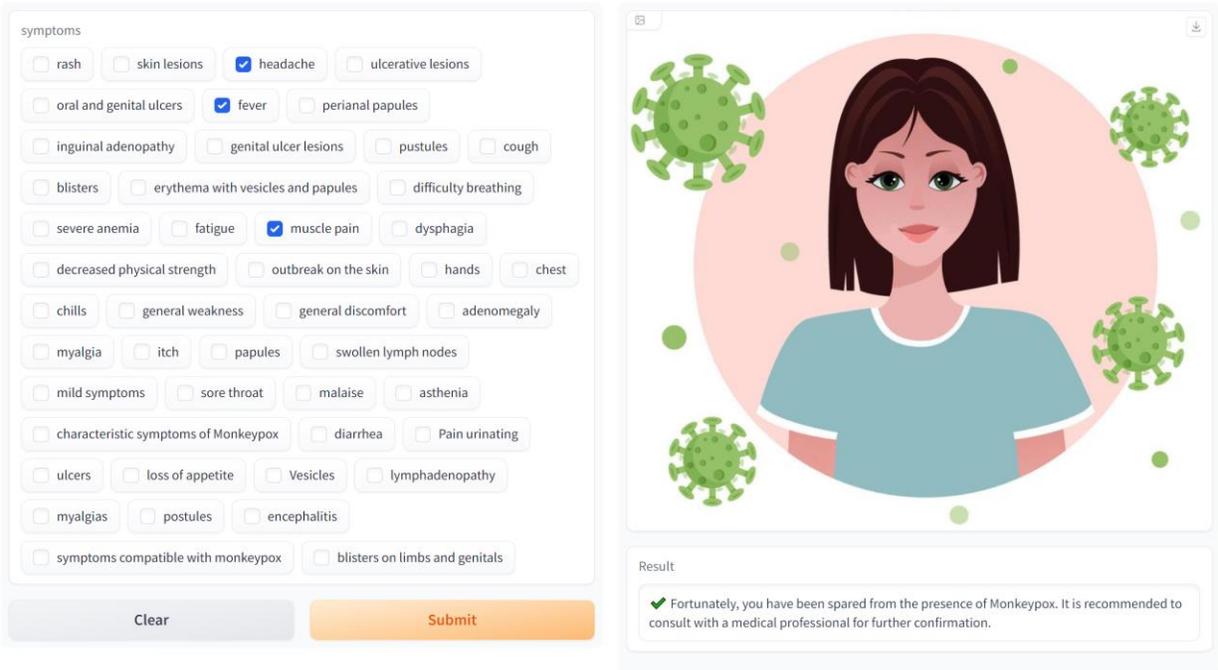

**Figure 3.** The result of SyMPox.

## 6. Impact overview

As previously discussed, this software can identify Monkeypox disease based on symptom analysis. The user-friendly interface ensures inclusivity across diverse social classes, enabling individuals from all backgrounds to input their symptoms and promptly determine the presence of Monkeypox signs. Additionally, the program operates remarkably quickly, allowing users to obtain results promptly.

The PCR is commonly employed for diagnosing this disease [3]. Despite the widespread availability of this test, its results may not be entirely conclusive. In contrast, SyMPox, renowned for its exceptional speed and accuracy, can be a valuable tool for healthcare providers in effectively detecting Monkeypox disease. This highlights the potential of SyMPox to offer enhanced diagnostic capabilities, surpassing the limitations associated with PCR testing.

As previously stated, this program eliminates the need for specialized equipment and can be easily utilized with a few straightforward interactions. Patients can conveniently input their symptoms through the user interface, and SyMPox subsequently performs classification to determine the presence or absence of Monkeypox signs in the individual.

## 7. Conclusion

This article presents SyMPox, an automated diagnostic system designed to identify Monkeypox using symptom-based analysis. With its emphasis on user-friendliness, exceptional accuracy, and widespread accessibility, this software can be effectively employed in various healthcare facilities, providing significant support to medical professionals. Developed using machine learning techniques, SyMPox achieved an impressive accuracy rate of 94.64%.

**Declaration of competing interest**

The authors declare that they have no known competing financial interests or personal relationships that could have appeared to influence the work reported in this paper.

**CRediT authorship contribution statement**

**A. Farzipour:** Methodology, Software, Validation, Investigation, Writing - Original Draft, Writing - Review & Editing, Visualization. **R. Elmi:** Methodology, Software, Validation, Investigation, Writing - Original Draft, Writing - Review & Editing, Visualization. **H. Nasiri:** Conceptualization, Methodology, Validation, Writing - Review & Editing, Supervision.